\documentclass{article}

\pdfoutput=1
\usepackage[utf8]{inputenc}
\usepackage[T1]{fontenc}
\usepackage{amsmath,amsfonts}
\usepackage{graphicx}
\usepackage{booktabs}
\usepackage{nicefrac}
\usepackage{microtype}
\usepackage{lipsum}
\usepackage{fancyhdr}
\usepackage{multirow}
\usepackage{algorithm}
\usepackage{algorithmic}
\usepackage{authblk}

\usepackage{PRIMEarxiv}
\usepackage{hyperref}

\graphicspath{{media/}}
\title{STFM: A Spatio-Temporal Information Fusion Model Based on Phase Space Reconstruction
 for Sea\\ Surface Temperature Prediction}

\author{
    Yin Wang$^{1,2}$\thanks{Corresponding author: \href{mailto:sduwangyin@163.com}{sduwangyin@163.com}}, 
     Chunlin Gong$^3$, 
     Xiang Wu$^4$, 
     Hanleran Zhang$^5$
  \\
    $^1$ School of Statistics and Mathematics, Shandong University of Finance and Economics, Jinan, Shandong, China \\
    $^2$ Jinan Fengdi Intelligent Electronics Co., Ltd., Jinan, Shandong, China \\
    $^3$ Shandong Zhike Intelligence Computing Co., Ltd., Jinan, Shandong, China \\
    $^4$ School of Mathematics and Physics, Anqing Normal University, Anqing, Anhui, China \\
    $^5$ SWUFE-UD Institute of Data Science, Southwestern University of Finance and Economics, Chengdu, Sichuan, China
}

\begin{document}
\maketitle

\begin{abstract}
The sea surface temperature (SST), a key environmental parameter, is crucial to optimizing production planning, making its accurate prediction a vital research topic. However, the inherent nonlinearity of the marine dynamic system presents significant challenges. Current forecasting methods mainly include physics-based numerical simulations and data-driven machine learning approaches. The former, while describing SST evolution through differential equations, suffers from high computational complexity and limited applicability, whereas the latter, despite its computational benefits, requires large datasets and faces interpretability challenges. This study presents a prediction framework based solely on data-driven techniques. Using phase space reconstruction, we construct initial-delay attractor pairs with a mathematical homeomorphism and design a Spatio-Temporal Fusion Mapping (STFM) to uncover their intrinsic connections. Unlike conventional models, our method captures SST dynamics efficiently through phase space reconstruction and achieves high prediction accuracy with minimal training data in comparative tests.\end{abstract}


\section{Introduction}
With the increasing exploration and development of the ocean, significant progress has been made in the study of ocean dynamic systems. Sea Surface Temperature (SST) is a key indicator of these complex systems, reflecting the heat exchange at the ocean surface and widely applied in marine dynamics research. SST variations directly impact various natural phenomena, which in turn have profound effects on human production and daily activities. Therefore, accurate prediction of SST is crucial for marine dynamics research and the development of ocean resources. However, due to the inherent strong nonlinearity of ocean activities, the ocean system is often treated as a chaotic system \cite{koshel2006chaotic}, making its behavior prediction extremely challenging.\\

Currently, two main approaches are used for SST prediction: physics-based numerical methods and data-driven methods. Chaos theory, a powerful tool for studying complex dynamic systems, shows that such systems may exhibit unpredictable behaviors. The Lorenz attractor \cite{lorenz1963deterministic} illustrates that system behavior can be described by deterministic equations, which theoretically enable short-term predictions based on initial conditions. The HCM numerical method \cite{barnett1993enso}, leveraging the coupling characteristics of the tropical ocean-atmosphere system, combines a nonlinear ocean circulation model with an SST-based statistical atmospheric model. Although this method represents the entire system using differential equations, the equations are often complex and computationally intensive. In contrast, data-driven methods predict system behavior by learning from historical data. With the rapid advancement of neural networks \cite{wu2018development}, time-series-based models such as Recurrent Neural Networks (RNN) \cite{rumelhart1986learning} and Long Short-Term Memory networks (LSTM) \cite{hochreiter1997long} have become widely used. Recently, LSTM-based SST prediction methods \cite{zhang2017prediction} have been proposed, which transform ocean sampling data into time series to capture temporal variations. To improve model robustness and sensitivity to outliers, Gaussian Process Regression (GPR) has been incorporated into LSTM models \cite{hittawe2022efficient}. \cite{fu2024prediction} combine the temporal processing ability of LSTM with Transformer networks \cite{vaswani2017attention} for more efficient data handling. Additionally, Variational Mode Decomposition (VMD) \cite{xu2023deep} has been introduced to decompose SST data into modes with different frequency characteristics, predicting each mode using the MIM model, and then combining them for high-accuracy SST predictions. Graph Neural Networks (GNN) have been employed to capture spatial correlations and temporal dynamics by considering sensor network topologies \cite{sun2021time}. The ConvLSTM model \cite{wanigasekara2024application}, which integrates Convolutional Neural Networks (CNN) for spatial feature extraction with LSTM for temporal feature extraction, has shown promising results in SST prediction. To better capture both spatial and temporal information, self-attention mechanisms \cite{yang2024attention} have been introduced to handle complex dynamic changes.\\

Previous data-driven studies have primarily used Convolutional Neural Networks (CNN) to extract spatial features when investigating the influence of spatial factors on SST. While CNNs can capture results after linear filtering of the temperature field, their ability to capture nonlinear relationships between spatial variables is limited. This study addresses this issue by proposing a time-series prediction model that integrates phase space reconstruction techniques with a Spatio-Temporal Fusion Mapping (STFM) module. Traditional phase space reconstruction methods build high-dimensional delayed attractors from low-dimensional initial attractors \cite{packard1980geometry}. Based on the Takens embedding theorem \cite{takens2006detecting}, we establish the Spatio-Temporal Information (STI) transformation equation for SST prediction \cite{chen2020predicting}. Furthermore, through the Spatio-Temporal Fusion module, the model learns the system’s dynamic features and establishes the mapping relationships on both sides of the equation. To assess the model’s performance over a broader spatial and temporal range, we adopt a parallelized SST prediction strategy in both time and space.\\

The SST dataset used in this study is sourced from the Optimized Interpolated Sea Surface Temperature (OISST) v2.1 dataset provided by the National Oceanic and Atmospheric Administration (NOAA). This dataset provides daily global SST records from September 1981, with a resolution of 1/4° latitude and longitude. The OISST dataset ensures high-accuracy temperature data by integrating satellite remote sensing and buoy observation data, using optimized interpolation methods. It has been widely used in climate research and marine dynamics. This dataset provides the SST data required for this study and is particularly suitable for validating our algorithmic model.\\

The main contributions of this study are as follows: 
\begin{itemize}
    \item The construction of the Spatio-Temporal Information (STI) transformation equation based on the Takens embedding theorem, which enhances the model’s interpretability and predictive performance.
    \item The design of the STFM module using a data-driven approach, requiring no boundary conditions or initial information, and relying on minimal training data.
    \item Extensive model performance experiments conducted across different marine regions, seasons, and prediction lengths, with ablation experiments assessing the contribution of each component to the overall model.
\end{itemize}

\section{Preliminary}
Since Sea Surface Temperature (SST) is a time series with chaotic characteristics, the use of embedding theorems and nonlinear analysis methods helps uncover the intrinsic relationships between spatial variables from historical data, thereby enabling more accurate predictions of future states.

\subsection*{2.1 Takens' Theorem}

Takens' embedding theorem \cite{takens2006detecting} is a powerful tool for studying chaotic time series. The theorem states the following:  

\textbf{Theorem 2.1:} For an infinite-length, noise-free scalar time series \( x(t) \) of a \( d \)-dimensional chaotic attractor, an \( m \)-dimensional embedding phase space can be found in a topologically invariant sense, where \( m \geq 2d + 1 \).

According to Takens' embedding theorem, we can reconstruct the corresponding high-dimensional phase space from a one-dimensional chaotic time series such that the reconstructed phase space is topologically equivalent to the original dynamical system.  
Typically, we use coordinate delay mapping to reconstruct the phase space, that is, constructing a phase space vector using the delay time \( \tau \) and embedding dimension \( m \):
\[
X(t) = [x(t), x(t + \tau), x(t + 2\tau), \dots, x(t + (m - 1)\tau)]
\]

where \( i = 0, 1, \dots, m-1 \), \( \tau \) is the time delay, and \( m \) is the embedding dimension. The reasonable choice of embedding dimension \( m \) and time delay \( \tau \) is crucial for the accuracy of phase space reconstruction. In general, \( \tau \) can be optimized using the mutual information method, while \( m \) should be determined based on the assumed attractor dimension. With this method, we can construct a phase space that is topologically equivalent to the original dynamical system, providing efficient data input features for subsequent time series forecasting.

\subsection*{2.2 Generalized Embedding Theorem}

Currently, although univariate time series reconstruction techniques are relatively mature, they have certain limitations in capturing the spatial characteristics of complex oceanic dynamical systems. The generalized embedding theorem proposed by Deyle et al. \cite{deyle2011generalized} provides a method that can reconstruct the attractor using multivariate time series. Compared to traditional univariate reconstruction methods, the generalized embedding theorem can consider the effects of multiple environmental or spatial factors on Sea Surface Temperature (SST) variations, thereby more effectively revealing the spatiotemporal dynamics of the system.  

Based on this theory, multivariate phase space vectors \( y_1, y_2, \dots, y_d \) can be constructed, where \( y_i(t) \) represents the state of the phase space vector \( y_i \) at time \( t \). To analyze the intrinsic relationships between spatial variables, the embedding vector of the multivariate time series is defined as follows:

\[
X(t) = 
\left[ 
\begin{array}{c}
y_1(t), \, y_2(t), \, \dots, \, y_d(t) ,\
y_1(t + (m-1)\tau), \\y_2(t + (m-1)\tau), \, \dots, \, y_d(t + (m-1)\tau)
\end{array}
\right]
\]

\subsection*{2.3 Spatial-Temporal Information Transformation Equation (STI)}

Based on the generalized embedding theorem, Chen \cite{chen2020predicting} proposed a method to reconstruct the attractor of a univariate time series using high-dimensional time series. The definition is as follows:

\textbf{Definition 2.1:} Let \( O \) represent the attractor observed by the system, and \( d_O \) represent the box-counting dimension of the attractor \( O \). The time series \( X(t_m) = [x_1(t_m), \dots, x_n(t_m)] \in \mathbb{R}^n \) represents the state of the system at time \( t_m \) on the attractor \( O \), where \( x_i(t_m) \) is the \( i \)-th component at time \( t_m \). The delay coordinate mapping \( \Psi: \mathbb{R}^n \to \mathbb{R}^L \) can be defined as:

\[
\Psi(X(t_m)) = [x_k(t_m), x_k(t_{m+1}), \dots, x_k(t_{m+L-1})] = Z(t_m)
\]

where \( x_k \) is the target variable of interest in the system, and the positive integer \( L \) satisfies \( L > 2d_O \). Based on the generalized embedding theorem, the reconstructed attractor \( Z(t_m) = [x_k(t_m), x_k(t_{m+1}), \dots, x_k(t_{m+L-1})] \in \mathbb{R}^L \) of the time series \( X(t_m) \) has a topological conjugacy relation with the non-delayed attractor \( O \), meaning that there exists a smooth injective function that associates the two.

For a known \( m \)-dimensional time series \( X(t_m) \) with \( M \) sampled points, the following mapping relation holds:

\[
\Psi(X(t_m)) = Z(t_m), \quad m = 1, 2, \dots, M
\]

Since \( Z(t_m) \) consists of \( L \) components of \( x_k \) at different time steps, the mapping \( \Psi \) can be written as a family of injective functions \( \Psi_1, \Psi_2, \dots, \Psi_L \). By constructing the following matrix form, we can represent the mapping process:

\[
\begin{aligned}
\left( \begin{matrix}
\Psi_1(X(t_1)) & \Psi_1(X(t_2)) & \cdots & \Psi_1(X(t_M)) \\
\Psi_2(X(t_1)) & \Psi_2(X(t_2)) & \cdots & \Psi_2(X(t_M)) \\
\vdots & \vdots & \ddots & \vdots \\
\Psi_L(X(t_1)) & \Psi_L(X(t_2)) & \cdots & \Psi_L(X(t_M))
\end{matrix} \right)
=
\left( \begin{matrix}
x_k(t_1) & x_k(t_2) & \cdots & x_k(t_M) \\
x_k(t_2) & x_k(t_3) & \cdots & x_k(t_{M+1}) \\
\vdots & \vdots & \ddots & \vdots \\
x_k(t_L) & x_k(t_{L+1}) & \cdots & x_k(t_{M+L-1})
\end{matrix} \right)
\end{aligned}
\]

\section{Methodology}
\subsection*{3.1 Problem Formula}

From the perspective of dynamical systems, the ocean temperature model is an infinite-dimensional dynamical system, which is continuous in both time and space. For computational convenience, we discretize the system. According to relevant research \cite{wang2022predicting}, we can efficiently reconstruct the dynamical system through sampling.

Let the initial time be \( t_0 \), and the sampling interval be \( \Delta t \). The \( i \)-th sampling time point \( t_i \) is given by:

\[
t_i = t_0 + i \cdot \Delta t
\]

Assuming that at any moment \( t \), we sample \( N \) points on the sea surface, we obtain the variables \( x_1, x_2, \dots, x_N \). At \( M \) different time points, we observe the sea surface temperature, and we can obtain the initial attractor \( O \).

It is important to note that the choice of sampling interval \( \Delta t \) is critical. Improper time delays may result in the reconstructed attractor failing to accurately represent the dynamical characteristics of the system.

Based on the above definitions, at the \( t_i \)-th sampling time, the temperature data for the sample point \( x_i \) is \( x_i(t_i) \), and the entire sampled sea area at \( t_i \) can be represented as \( X(t_i) \), where \( X(t_i) = [x_1(t_i), x_2(t_i), \dots, x_N(t_i)] \). Thus, the initial attractor \( O \) constructed from \( M \) samples can be represented as:

\[
X \in \mathbb{R}^{M \times N} = \left[ X(t_1), X(t_2), \dots, X(t_M) \right]^T
\]

\begin{figure}[h]
  \centering
  \includegraphics[width=7in]{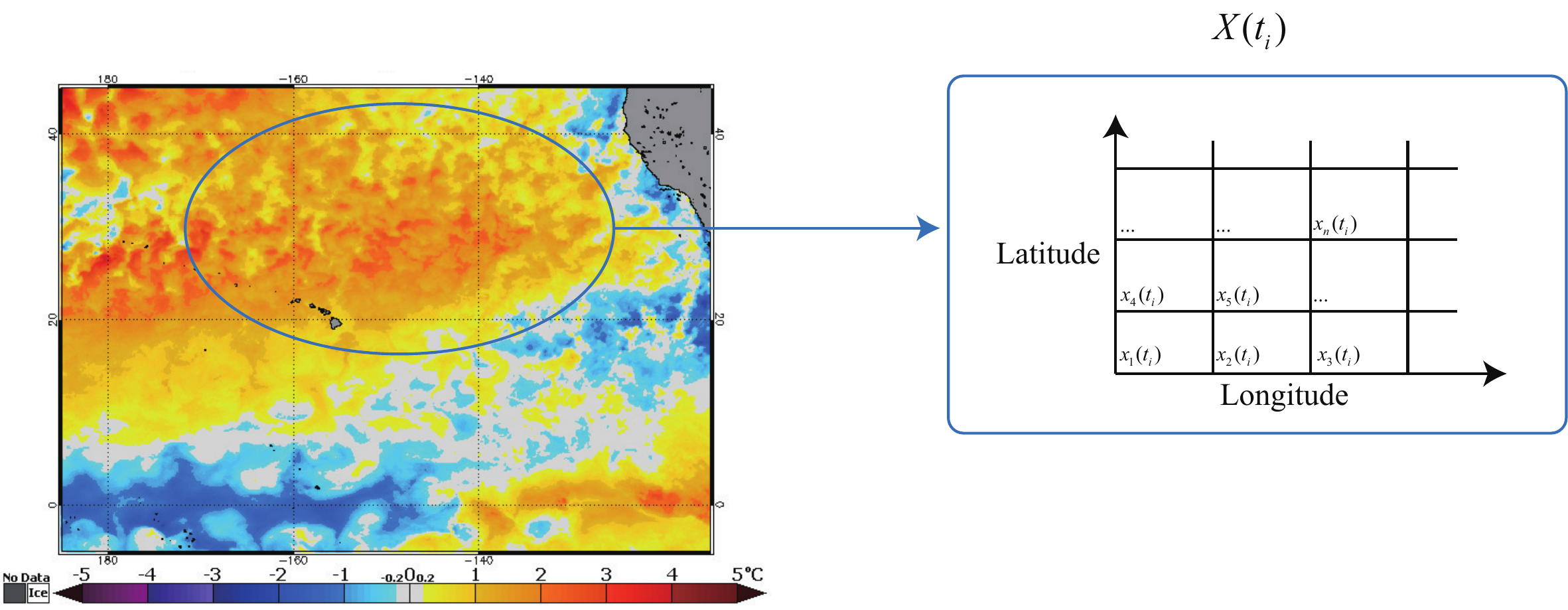}  
  \caption{Discrete sampling of sea surface temperature.} 
  \label{fig:1} 
\end{figure}

As shown in \autoref{fig:1}

If we perform an \( L \)-step prediction on the variable \( x_k \), we get:

\[
\hat{x}_k^L = \left[ \hat{x}_k(t_{M+1}), \hat{x}_k(t_{M+2}), \dots, \hat{x}_k(t_{M+L}) \right]
\]

To capture the spatiotemporal and nonlinear features of the high-dimensional dynamical system, we represent the delayed attractor as a matrix \( D \):

\[
D = \left( \begin{matrix}
   x_k(t_1) & x_k(t_2) & \cdots & x_k(t_M)  \\
   x_k(t_2) & x_k(t_3) & \cdots & x_k(t_{M+1})  \\
   x_k(t_3) & x_k(t_4) & \cdots & x_k(t_{M+2})  \\
   \vdots & \vdots & {} & \vdots   \\
   x_k(t_{L+1}) & x_k(t_{L+2}) & \cdots & x_k(t_{M+L})  \\
\end{matrix} \right)
\]

From \textbf{Theorem 2.1}, if the box dimension of the initial attractor \( O \) is \( d \), and the box dimension of the delayed attractor \( D \) is \( m \), then if \( m \geq 2d + 1 \), there exists a smooth mapping \( \Psi \) such that the original attractor \( O \) is topologically equivalent to the delayed embedded attractor \( D \). This mapping \( \Psi: O \to D \) can reconstruct the system's nonlinear dynamical characteristics. By choosing an appropriate \( \Psi \), we can map the initial attractor \( O \) to the delayed attractor \( D \), thus obtaining the Spatiotemporal Information Transformation Equation (STI):

\[
O = \left( \begin{matrix}
   x_1(t_1) & x_2(t_1) & \cdots & x_N(t_1)  \\
   x_1(t_2) & x_2(t_2) & \cdots & x_N(t_2)  \\
   \vdots & \vdots & {} & \vdots   \\
   x_1(t_M) & x_2(t_M) & \cdots & x_N(t_M)  \\
\end{matrix} \right)
\]
\[
\xrightarrow{\Psi}
\]
\[
D = \left( \begin{matrix}
   x_k(t_1) & x_k(t_2) & \cdots & x_k(t_M)  \\
   x_k(t_2) & x_k(t_3) & \cdots & x_k(t_{M+1})  \\
   x_k(t_3) & x_k(t_4) & \cdots & x_k(t_{M+2})  \\
   \vdots & \vdots & {} & \vdots   \\
   x_k(t_L) & x_k(t_{L+1}) & \cdots & x_k(t_{M+L-1})  \\
\end{matrix} \right)
\]

\begin{figure}[h]
  \centering
  \includegraphics[width=3in]{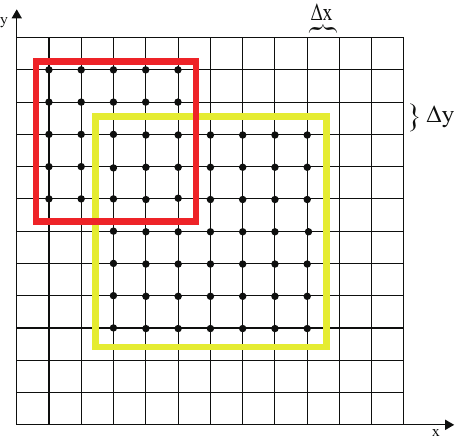}  
  \caption{Two different variable association options. The yellow rectangle represents the area to be predicted. Within this region, for any fixed grid point, the red sliding rectangle centered at the target variable location contains the grid points to be associated with the variable.} 
  \label{fig:2}  
\end{figure}

Based on the Spatiotemporal Information Transformation Equation (STI), we parallelize it spatially, obtaining the spatially parallel STI equation as:

\[
\Psi(O_k) = D_k, \quad k = 1, 2, \dots, D_x D_y
\]

where \( D \) represents the total number of grid points within the variable association region.Spatially parallel sampling as in \autoref{fig:2}
\subsection*{3.2 Model Architecture}

Based on the definitions in Section 3.1, this paper uses a neural network to learn the nonlinear characteristics of the mapping \( \Psi \), to achieve the transformation from the initial attractor \( O \) to the delayed attractor \( D \). If the initial attractor \( O \) is taken as the input and the delayed attractor \( D \) as the output, the mapping relationship can be expressed as:

\[
\begin{aligned}
  & \Psi(O) = D \\ 
  & O \in \mathbb{R}^{M \times N}, \quad D \in \mathbb{R}^{L \times M}
\end{aligned}
\]

Using spatiotemporal information transformation, the smooth mapping \( \Psi \) is decomposed into a series of single-step prediction functions, that is, \( L \) single-step prediction functions \( \{{\Psi }_{1}, {\Psi }_{2}, \dots, {\Psi }_{L}\} \). This is defined as:

\[
{\Psi }_{i}\left( X\left( {t}_{k} \right) \right)={x}_{k}\left( {t}_{k+i-1} \right)
\]

Through this step, the complex nonlinear relationships in the multivariate time series are split into multiple single-step prediction tasks, which reduces the computational complexity of the model and improves prediction accuracy. The matrix form of the single-step prediction functions is given by:

\[
\left( \begin{matrix}
   {\Psi }_{1}(X({t}_{1})) & {\Psi }_{1}(X({t}_{2})) & \cdots  & {\Psi }_{1}(X({t}_{M}))  \\
   {\Psi }_{2}(X({t}_{1})) & {\Psi }_{2}(X({t}_{2})) & \cdots  & {\Psi }_{2}(X({t}_{M}))  \\
   \vdots  & \vdots  & {} & \vdots   \\
   {\Psi }_{L}(X({t}_{1})) & {\Psi }_{L}(X({t}_{2})) & \cdots  & {\Psi }_{L}(X({t}_{M}))  \\
\end{matrix} \right)\]
\[=\left( \begin{matrix}
   {x}_{k}({t}_{1}) & {x}_{k}({t}_{2}) & \cdots  & {x}_{k}({t}_{M})  \\
   {x}_{k}({t}_{2}) & {x}_{k}({t}_{3}) & \cdots  & {x}_{k}({t}_{M+1})  \\
   \vdots  & \vdots  & {} & \vdots   \\
   {x}_{k}({t}_{L}) & {x}_{k}({t}_{L+1}) & \cdots  & {x}_{k}({t}_{M+L-1})  \\
\end{matrix} \right)
\]

From the above equation, we can see that to establish the nonlinear relationship between the original attractor and the delayed attractor, we need to solve for \( L \) different nonlinear mappings. However, due to the inherent advantage of neural networks in handling high-dimensional data, when designing the learning algorithm, we can unify these \( L \) mappings into a single mapping \( \Psi \). Next, we construct an appropriate neural network model to fit the nonlinear mapping \( \Psi \). 

\begin{figure}[h]
  \centering
  \includegraphics[width=7in]{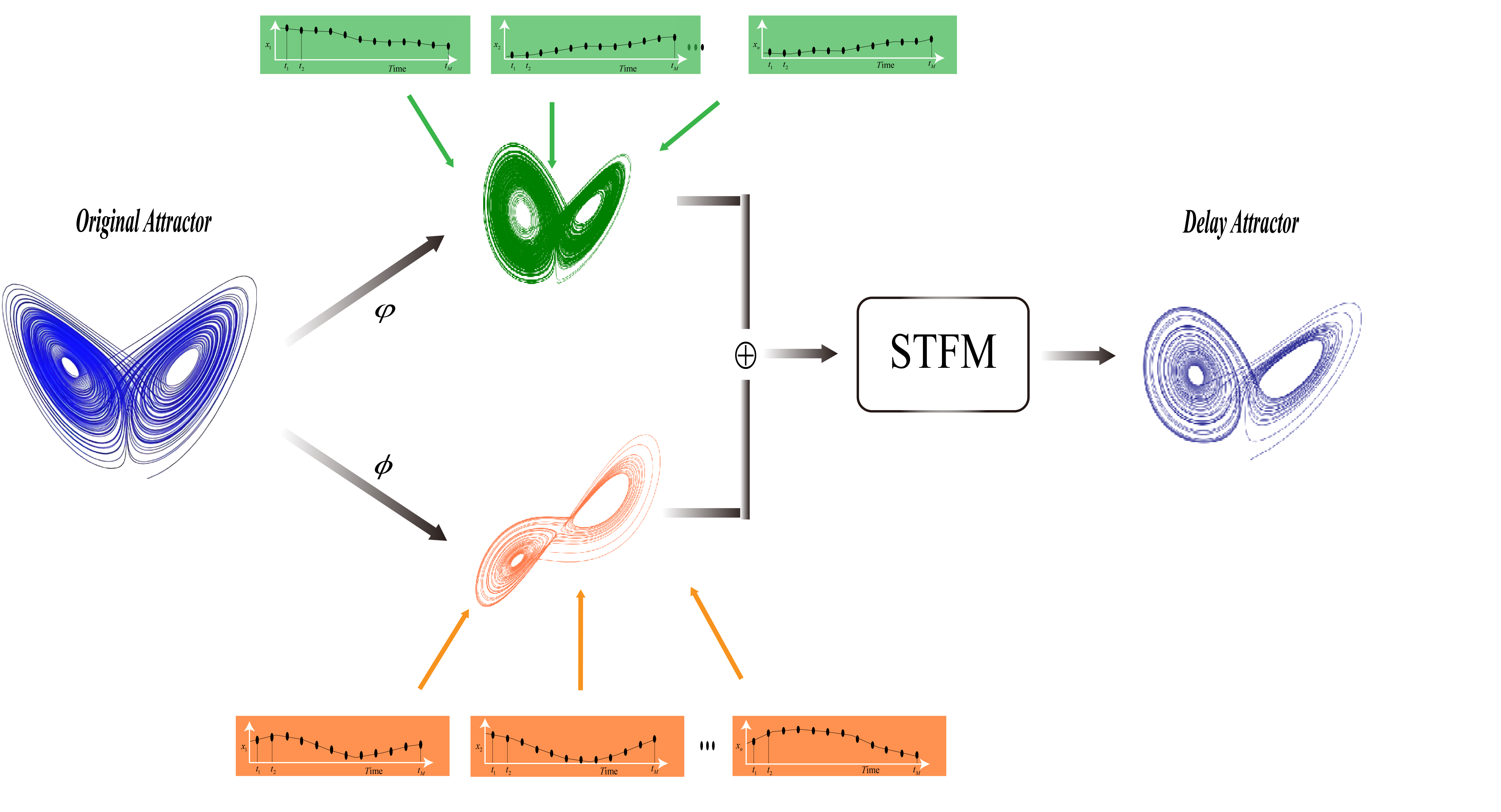}  
  \caption{The overall model architecture is shown, including the layout of the seasonal and trend mapping modules, and the process of obtaining delayed attractors through the time-space aggregation module.} 
  \label{fig:3} 
\end{figure}

To capture the seasonal and long-term trends of temperature changes at each spatial point, the model utilizes two modules (as shown in \autoref{fig:3}) to extract the seasonal and long-term trend information from the time series. Based on this, the seasonal and long-term trend information is aggregated through the Spatiotemporal Feature Aggregation Module (STFM) to adjust the embedding dimension, resulting in the delayed attractor \( D \).

The spatial variables in the initial attractor are:

\[
x_{i}^{M}=\left[ {x}_{i}({t}_{1}),{x}_{i}({t}_{2}),\dots,{x}_{i}({t}_{M}) \right]
\]

By using the trend mapping, the long-term trend prediction information \( {G}_{i} \) for all univariate attractors in the system is extracted. Similarly, the seasonal prediction information \( {F}_{i} \) is extracted using the seasonal mapping \( \phi \), as follows:

\[
\begin{aligned}
  \varphi (X^{T} \in \mathbb{R}^{N\times M}) &= G \in \mathbb{R}^{N\times M} \\
  \phi (X^{T} \in \mathbb{R}^{N\times M}) &= F \in \mathbb{R}^{N\times M}
\end{aligned}
\]

For the extraction of seasonal and trend information of the univariate initial attractor, we can add the matrices \( F_{i} \) and \( G_{i} \) to obtain:

\[
\hat{x}_{i}^{M}=\left[ {x}_{i}({t}_{M+1}), {x}_{i}({t}_{M+2}), \dots, {x}_{i}({t}_{2M}) \right]
\]

Using the trend mapping \( \varphi \) and seasonal prediction \( \phi \), we obtain the intermediate variable:

\[
{{\hat{X}}^{M}} = \left[ \hat{x}_{1}^{M}, \hat{x}_{2}^{M}, \dots, \hat{x}_{N}^{M} \right]^T
\]

Further, we input the intermediate variable \( {\hat{X}}^{M} \) into the Spatiotemporal Feature Aggregation Module (STFM) for time-space feature extraction and dimension adjustment, resulting in:

\[
D = \Phi \left( {{\hat{X}}^{M}} \right)
\]

Based on the above description, we can derive the nonlinear fitting mapping \( \Psi \) as:

\[
\Psi \left( O \right) = STFM \left( {{\hat{X}}^{M}} \right) = STFM(\varphi(X^{T}) + \phi(X^{T})) = D
\]

Next, we will detail the design of each module.

\subsection*{3.2.1 Trend Module}

The design of the trend mapping module aims to extract the variation trend of the univariate initial attractor \( x_{i}^{M} = \left[ x_{i}(t_{1}), x_{i}(t_{2}), \dots, x_{i}(t_{M}) \right] \) at different time scales.

The goal of the trend function \( \varphi \) is to capture the long-term trend change of the initial attractor and obtain the trend prediction data. Let the input be \( X^{T} \), and the trend mapping function can be expressed as:

\[
\varphi \left( X^{T} \in \mathbb{R}^{N \times M} \right) = F \in \mathbb{R}^{N \times M}
\]

Since the trend mapping mainly reflects the trend information of the same variable over different times, to further improve the predictive ability of the model, \( \varphi \) can be decomposed into a series of single-step mapping functions \( \left[ \varphi_1, \varphi_2, \dots, \varphi_N \right] \). The mapping definition for \( \varphi_i \) is:

\[
\varphi_{i}\left( x_{i}^{M} \in \mathbb{R}^{1 \times M} \right) = F_{i} \in \mathbb{R}^{1 \times M}
\]

\begin{figure}[h]
  \centering
  \includegraphics[width=5in]{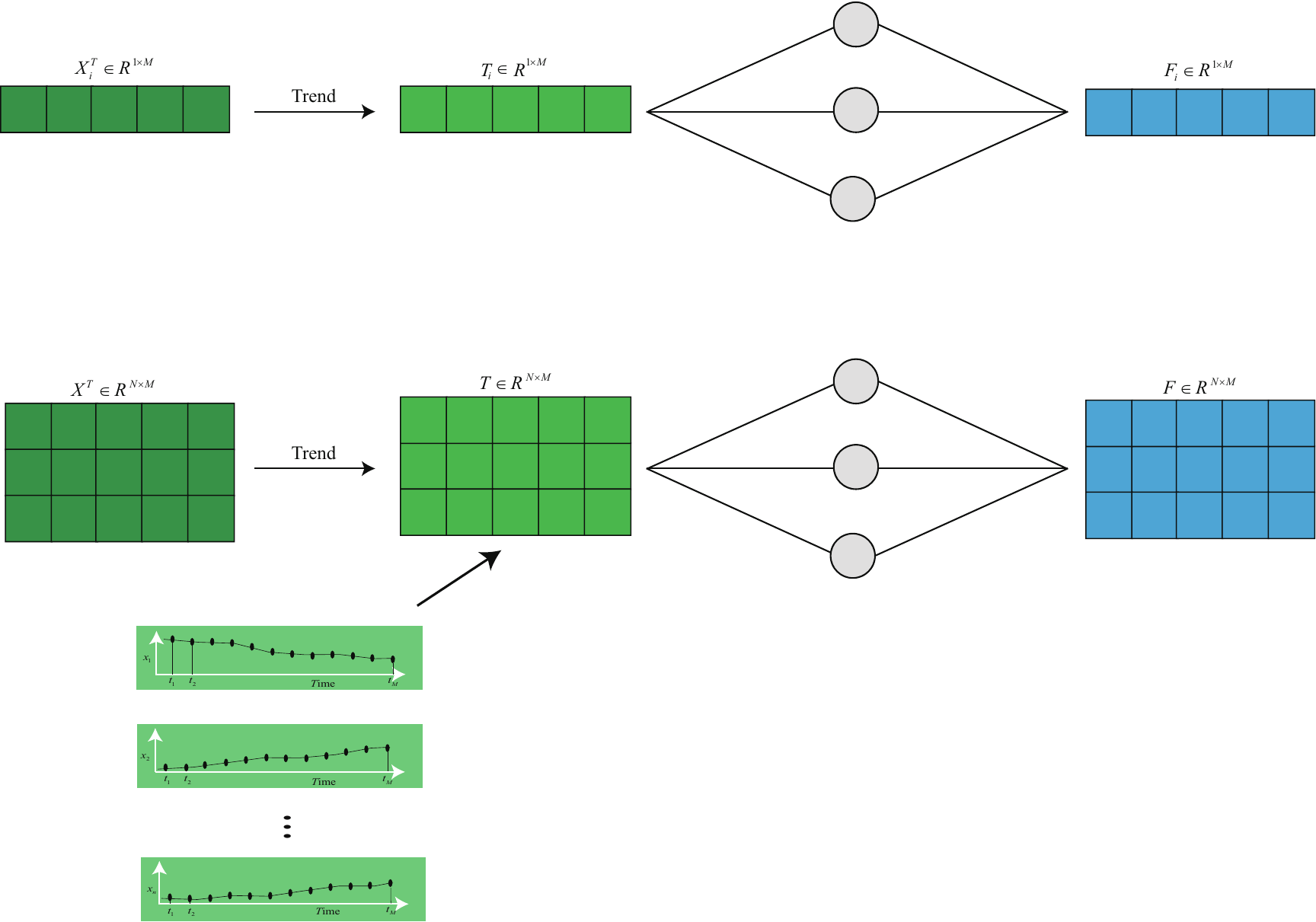} 
  \caption{Demonstrates the overall architecture of the single-step mapping \( \varphi \) for the initial attractor \( O \). First, the input data is processed using the sliding window method to extract local trend features, and the trend decomposition module \( \text{Trend} \) extracts the trend information \( T \) from the time series. The obtained trend features are then processed through multiple layers of fully connected layers to capture long-term dependencies, ultimately generating the trend prediction matrix \( F \).}  
  \label{fig:4}  
\end{figure}

As shown in \autoref{fig:4}, the trend function \( \varphi \) consists of the trend decomposition module and a fully connected layer. Define the trend data as \( T \). For each variable \( x_{i}(t_{j}) \), we define its corresponding trend variable as \( T_{i}(t_{j}) \). In this process, we use the sliding window method to capture the dynamic trend of the time series. Let the current window size be \( \text{Size} \), and the trend variable \( T_{i}(t_{j}) \) is computed as:

\[
T_{i}(t_{j}) \xrightarrow{\text{Trend}} \left\{ 
\begin{array}{l}
\frac{1}{j+1} \sum_{k=1}^{j} x_{i}(t_{k}), \quad \text{if } j < \text{Size} \\
\frac{1}{\text{Size}} \sum_{k=j-\text{Size}+1}^{j} x_{i}(t_{k}), \quad \text{if } j \geq \text{Size}
\end{array}
\right.
\]

Based on the trend mapping function, we obtain the trend data for the univariate variable \( i \) as \( T_{i} = [T_{i}(t_{1}), T_{i}(t_{2}), \dots, T_{i}(t_{M})] \). By combining the trend data for all variables into a matrix form, we have:

\[
T = \left( \left[ T_{1}, T_{2}, \dots, T_{N} \right] \right)^{T}
\]

After constructing the trend matrix \( T \), we input it into the fully connected layer to extract the trend information, ultimately generating the trend prediction matrix \( F \in \mathbb{R}^{N \times M} \), as shown in the following equation:

\[
F = \text{Linear}(T \in \mathbb{R}^{N \times M}) = \text{Trend}(X^{T} \in \mathbb{R}^{N \times M})
\]

\subsection*{3.2.2 Seasonal Module}

Define the seasonal mapping function as \( \phi \). This mapping function can be used to compute the relationship between the initial attractor and the seasonal prediction matrix \( G \) of the time series:

\[
\phi \left( X^{T} \in \mathbb{R}^{N \times M} \right) = G \in \mathbb{R}^{N \times M}
\]

If the mapping \( \phi \) is decomposed into a family of injective functions \( \left[ \phi_{1}, \phi_{2}, \dots, \phi_{N} \right] \), where each \( \phi_{i} \) is considered as the single-step mapping prediction for the univariate initial attractor \( x_{i}^{M} = \left[ x_{i}(t_{1}), x_{i}(t_{2}), \dots, x_{i}(t_{M}) \right] \), then the mapping \( \phi_{i} \) is defined as:

\[
\phi_{i} \left( x_{i}^{M} \in \mathbb{R}^{1 \times M} \right) = G_{i} \in \mathbb{R}^{1 \times M}
\]

\begin{figure}[h]
  \centering
  \includegraphics[width=5in]{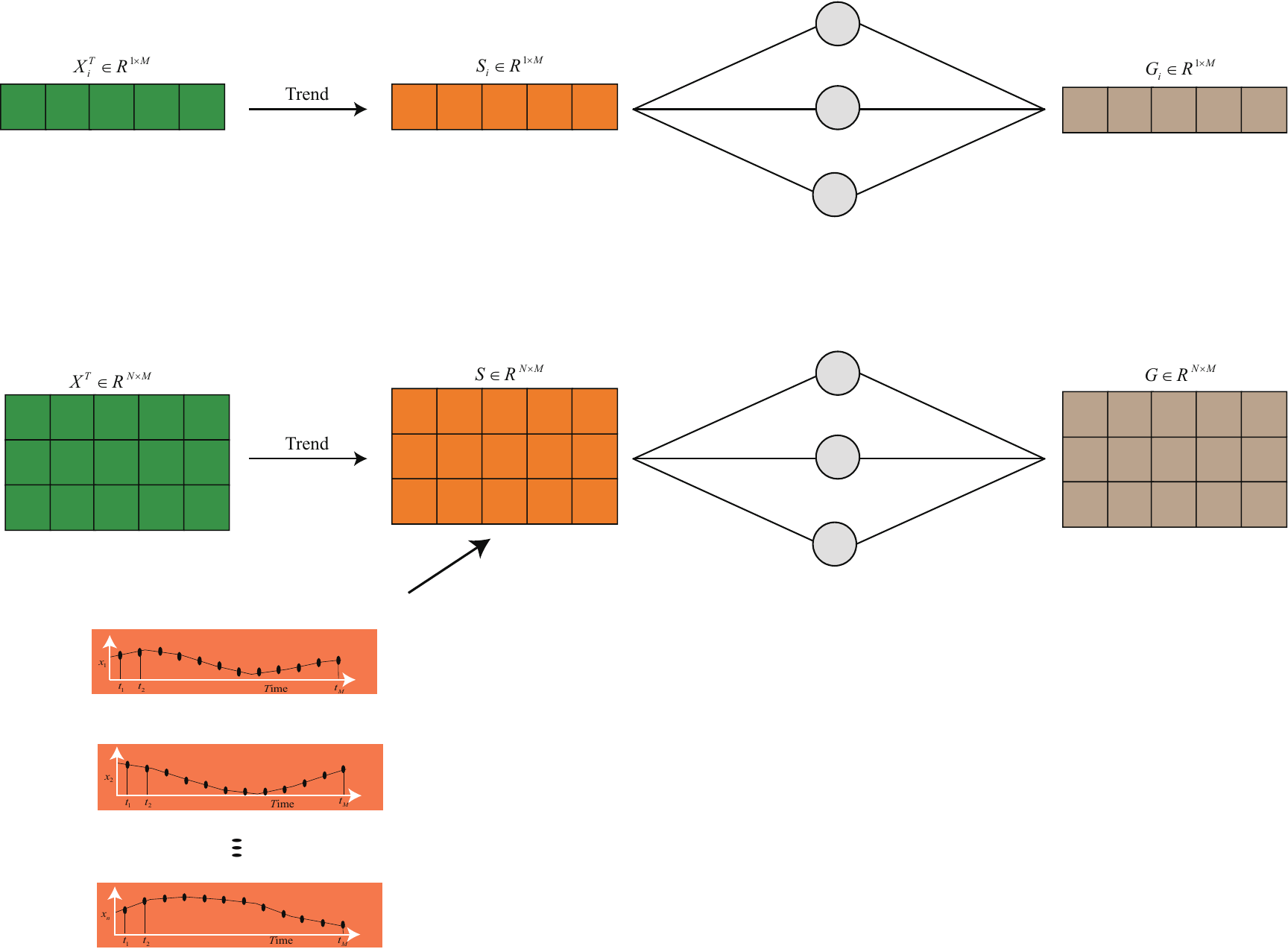} 
  \caption{Demonstrates the single-step mapping for the univariate initial attractor \( x_{i}^{M} = \left[ x_{i}(t_{1}), x_{i}(t_{2}), \dots, x_{i}(t_{M}) \right] \) and the overall mapping structure of \( X^{T} \). First, the input data is processed using the sliding window method to extract local seasonal features, and the seasonal decomposition module \( \text{Season} \) extracts the seasonal information \( S \) from the time series. The obtained seasonal features are then processed through multiple layers of fully connected layers to further mine the periodic variations in the seasonal features, ultimately generating the seasonal prediction matrix \( G \).}  
  \label{fig:5}  
\end{figure}

As shown in \autoref{fig:5}, the mapping function includes the seasonal decomposition module \( \phi \) and a fully connected layer. Similar to the trend data extraction method, define the seasonal data as \( S \). For each variable \( x_{i}(t_{j}) \), we define its corresponding seasonal variable as \( S_{i}(t_{j}) \). In this process, we use the sliding window method to capture the dynamic trend of the time series. Let the current window size be \( \text{Size} \), and the seasonal variable \( S_{i}(t_{j}) \) is computed as:

\[
S_{i}(t_{j}) = x_{i}(t_{j}) - T_{i}(t_{j})
\]

A more general expression is:

\[
S_{i}(t_{j}) \xrightarrow{\text{Season}} \left\{ 
\begin{array}{l}
x_{i}(t_{j}) - \frac{1}{j+1} \sum_{k=1}^{j} x_{i}(t_{k}), j < \text{Size} \\
x_{i}(t_{j}) - \frac{1}{\text{Size}} \sum_{k=j-\text{Size}+1}^{j} x_{i}(t_{k}),  j \geq \text{Size}
\end{array}
\right.
\]

Using the above formula, the mapping function \( \text{Season} \) can decompose the time series of each univariate variable \( x_{i} \) into seasonal data \( S_{i} = [S_{i}(t_{1}), S_{i}(t_{2}), \dots, S_{i}(t_{M})] \). By combining the seasonal data of all variables at each time point, we form a matrix \( S \), expressed as:

\[
\mathbf{S} = \left[ 
\begin{matrix}
   S_{1}(t_{1}) & S_{1}(t_{2}) & \dots  & S_{1}(t_{n})  \\
   S_{2}(t_{1}) & S_{2}(t_{2}) & \dots  & S_{2}(t_{n})  \\
   \vdots  & \vdots  & \ddots  & \vdots   \\
   S_{m}(t_{1}) & S_{m}(t_{2}) & \dots  & S_{m}(t_{n})  \\
\end{matrix} 
\right]
\]

Once the seasonal information matrix \( S \) is constructed, in order to better capture the periodic patterns in the data, we input the matrix \( S \) into the fully connected layer, resulting in the predicted seasonal matrix \( G \in \mathbb{R}^{N \times M} \), expressed as:

\[
G = \text{Linear}(S \in \mathbb{R}^{N \times M}) = \text{Season}(X^{T} \in \mathbb{R}^{N \times M})
\]

\subsection*{3.2.3 Original Spatio-Temporal Fusion Module (STFM)}

\begin{figure}[h]
  \centering
  \includegraphics[width=7in]{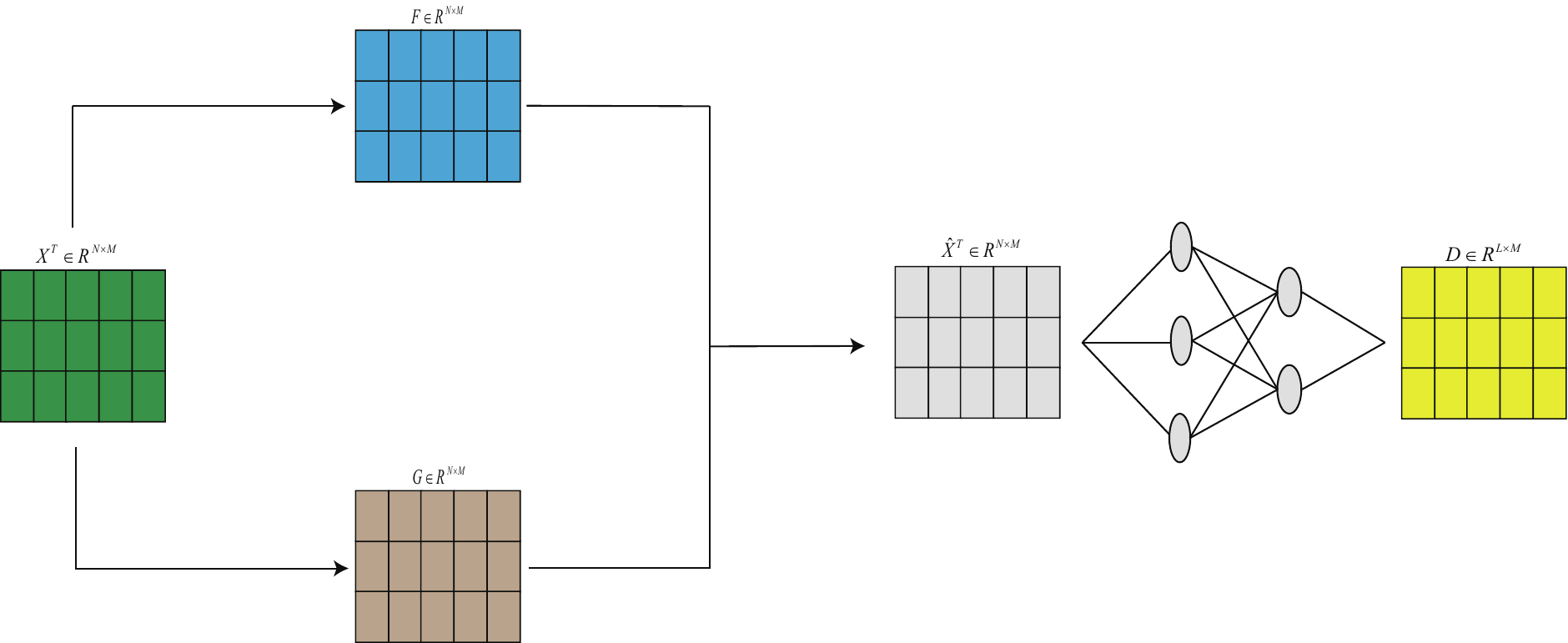} 
  \caption{Demonstrates the full process of the initial model. After obtaining the predicted matrix \( \hat{X}^{M} \), we used a simple two-layer fully connected neural network to obtain the final delayed attractor matrix.} 
  \label{fig:6} 
\end{figure}

When the model was initially formulated, we attempted to adjust the dimensional matching and information aggregation between the predicted matrix \( \hat{x}_{i}^{M} = \left[ x_{i}(t_{M+1}), x_{i}(t_{M+1}), \dots, x_{i}(t_{2M}) \right] \in \mathbb{R}^{N \times M} \) and the delayed attractor \( D \in \mathbb{R}^{L \times M} \) using a two-layer fully connected neural network \autoref{fig:6}. That is:

\[
\begin{aligned}
  D &= W_2 \cdot \text{ReLU}(W_1 \cdot \hat{x}_{i}^{M} + b_1) + b_2 \\
  W&_1, \in \mathbb{R}^{\text{Hidden} \times N}, \quad W_2 \in \mathbb{R}^{L \times \text{Hidden}}
\end{aligned}
\]

However, this simple construction led to several drawbacks. First, the limited number of network layers restricted the model's expressive power, preventing it from effectively capturing the complex patterns in the dynamical system. Therefore, even though underfitting did not occur, the model still performed poorly on the training set, and the RMSE value remained relatively high on the test set. Secondly, the limited number of network layers caused instability during the training process. With the same training and testing sets, the model's test results fluctuated greatly, indicating that during training, the variance of the data was highly influenced by noise, making the model excessively sensitive to noise and leading to significant fluctuations.

\subsection*{3.2.4 Enhanced Spatio-Temporal Fusion Module (STFM-V1)}

\begin{figure}[htbp]
  \centering
  \includegraphics[width=4in]{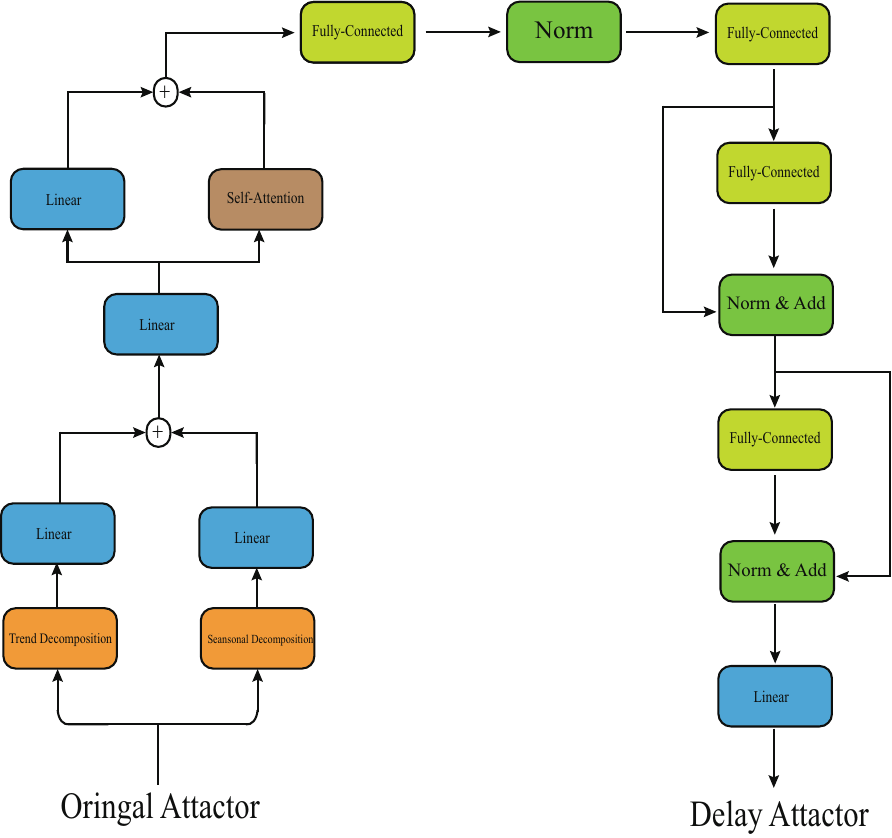} 
  \caption{Demonstrates the single-step mapping for the univariate initial attractor \( x_{i}^{M} = \left[ x_{i}(t_{1}), x_{i}(t_{2}), \dots, x_{i}(t_{M}) \right] \) and the overall mapping structure of \( X^{T} \). First, the input data is processed using the sliding window method to extract local seasonal features, and the seasonal decomposition module \( \text{Season} \) extracts the seasonal information \( S \) from the time series. The obtained seasonal features are then processed through multiple layers of fully connected layers to further mine the periodic variations in the seasonal features, ultimately generating the seasonal prediction matrix \( G \).}  
  \label{fig:7}  
\end{figure}

The STFM-V1 model is an important adjustment to the original structure to enhance its performance and stability, as shown in \autoref{fig:7}. To increase the correlation between the data, the model introduces the Self-Attention module \cite{yang2024attention}, which strengthens the interactions and dependencies between different data features, thereby improving the model's ability to capture complex patterns. Additionally, a normalization layer \cite{ba2016layer} has been added. Given the increase in the number of network layers, deep networks may face the problem of vanishing gradients. To address this, residual connections \cite{he2016deep} are added after the normalization layer, preventing the vanishing gradient phenomenon and enhancing the stability and convergence speed of the training process.

During the training process, it was observed that the predicted value \( x_k(t_{M+1}) \) often had a large deviation from the true value, which might be due to the model's failure to find a suitable baseline. To solve this issue, before training starts, the model adds the value at the last time point of the training set, \( x_k(t_{M}) \), to the initial attractor matrix as a reference baseline. To avoid shifting the predicted values overall, the model is guided to be closer to the true value during the initial prediction by adjusting the hyperparameter \( \lambda_{\text{base}} \). The corresponding formula is as follows:

\[
\text{loss}(\Psi'(O), D) = \text{MSE}(\Psi(O), D) + \lambda_{\text{base}} x_k(t_M)
\]

The construction of the initial attractor with the reference baseline \(O\)' is as follows:

\[
\left[\begin{matrix}
   x_1(t_1) + \lambda_{\text{base}} \times x_k(t_M) & \cdots  & x_N(t_1) + \lambda_{\text{base}} \times x_k(t_M)  \\
   x_1(t_2) + \lambda_{\text{base}} \times x_k(t_M) & \cdots  & x_N(t_2) + \lambda_{\text{base}} \times x_k(t_M)  \\
   \vdots  & \vdots  & \vdots  &  \\
   x_1(t_M) + \lambda_{\text{base}} \times x_k(t_M) & \cdots  & x_N(t_M) + \lambda_{\text{base}} \times x_k(t_M)  \\ 
\end{matrix} \right]
\]

During the training process, since the delayed attractor is constructed as:

\[
D = \left( \begin{matrix}
   x_k(t_1) & x_k(t_2) & \cdots  & x_k(t_M)  \\
   x_k(t_2) & x_k(t_3) & \cdots  & x_k(t_{M+1})  \\
   \vdots  & \vdots  & {} & \vdots   \\
   x_k(t_L) & x_k(t_{L+1}) & \cdots  & x_k(t_{M+L-1})  \\
\end{matrix} \right)
\]

Based on the construction principle of the delayed attractor, the corresponding values at the same time point in the delayed embedding matrix (i.e., the elements on the reverse diagonal) should satisfy the consistency constraint. To implement this constraint, we modify the loss function: First, we extract all the reverse diagonal elements from the delayed attractor matrix and construct a sequence. Then, we calculate the variance of this sequence as a regularization term, denoted as \( \text{diagonal\_loss} \). By introducing the hyperparameter \( \lambda_{\text{diag}} \) to adjust the regularization weight, we integrate this term into the loss function, forcing the model to satisfy the time consistency condition during training and ensuring that the geometric structure of the delayed attractor is preserved. The mathematical expression is:

\[
\text{loss}(\Psi'(O'), D) = \text{MSE}(\Psi(O) + \lambda_{\text{base}}  x_k(t_M), D)\] 
\[+ \lambda_{\text{diag}}  \text{diagonal\_loss}
\]

\section{Experiment}
\subsection*{4.1 Parametric Experiment}

Considering that NOAA provides ample SST data, we select the variable association method shown by the red rectangle in Figure 2. The center sampling point for the experiment is located at 10.125°N, 115.125°E in the central region of the South China Sea. Unless otherwise stated, the sampling start time for the experiment is October 1, 2013, with predictions made every 5 days. The forecast length \( L = 15 \) days, and the sampling length \( M = 30 \) days.

Our model method is as shown in \autoref{alg:STFM}.

To investigate the impact of spatial dependency region scale on prediction performance, we systematically tested the model's performance under different spatial ranges. As shown in Table I, the root mean square error (RMSE) decreases monotonically as the region scale increases. This phenomenon can be attributed to the larger spatial domain providing richer dynamical system information, thereby improving the reconstruction accuracy of the initial attractor.

Comparative analysis shows that the STFM model demonstrates stable prediction performance in small-scale regions. However, as the region scale increases, its RMSE significantly rises and fluctuates, indicating limitations in regional feature extraction capability. In contrast, STFM-V1 demonstrates excellent scalability as the spatial domain expands, reducing its RMSE by 7\% compared to the persistent model, validating the effectiveness of the improved regional feature learning mechanism.

\begin{table}[H]
\centering
\large 
\setlength{\tabcolsep}{3pt}
\renewcommand{\arraystretch}{1.25}
\begin{tabular}{ccccccc}
\hline
& \textbf{$\Delta$lat $\times$ $\Delta$lon}  &1$^{\circ} \times$1$^{\circ}$ & 2$^{\circ} \times$2$^{\circ}$ & 3$^{\circ} \times$3$^{\circ}$  \\
\hline
\multirow{2}{*}{\textbf{STFM}} & RMSE & 1.5486 & 1.4040 & 3.4419  \\
                                 & MAPE & 5.3922 & 4.8612 & 12.1309  \\
\hline
\multirow{2}{*}{\textbf{STFM-V1}}   & RMSE & 0.7059 & 0.6939 & 0.6783  \\
                                 & MAPE & 2.3482 & 2.3059 & 2.2537  \\
\hline
\multirow{2}{*}{\textbf{Persistent}}& RMSE & 1.1537 & 1.1537 & 1.1537  \\
                                 & MAPE & 3.9751 & 3.9751 & 3.9751  \\
\hline
\end{tabular}
\caption{ IMPACT OF DIFFERENT REGIONAL SCALES ON PREDICTION ACCURACY}
\end{table}

Table II presents the prediction performance under different time intervals. We set \( \Delta t = 1, 3, 5 \) and compared our two models with LSTM, XGBoost, DNN, and the Persistent Model. The comparison shows that STFM-V1 performs better at larger time intervals, with a significant advantage over the other models.

Further, with a fixed time interval \( \Delta t = 5 \) and a sampling time length of \( M = 30 \), we evaluated the model's one-step prediction (\( L = 2 \)) and multi-step prediction (\( L > 2 \)) performance. As shown in Table III, our STFM-V1 model is more stable than the other models, with prediction accuracy less affected by the prediction length. Moreover, as the prediction length increases, the STFM-V1 model demonstrates lower prediction errors, indicating better long-term prediction capability compared to traditional methods.

\begin{table}[h]
\centering
\large 
\setlength{\tabcolsep}{6pt} 
\renewcommand{\arraystretch}{1.25}
\begin{tabular}{ccccccc}
\hline
& \textbf{$\Delta$t}  & 1 & 3 & 5  \\
\hline
\multirow{2}{*}{\textbf{STFM}} & RMSE & 0.9476 & 1.8787 & 1.1887 \\
                                 & MAPE & 3.3535 & 6.3528 & 4.0798 \\
\hline
\multirow{2}{*}{\textbf{STFM-V1}}   & RMSE & 0.9210 & 0.8139 & 0.6939 \\
                                 & MAPE & 3.2524 & 2.7719 & 2.3059 \\
\hline
\multirow{2}{*}{\textbf{LSTM}}& RMSE & 1.3513 & 2.1198 & 1.8597 \\
                                 & MAPE & 4.7904 & 7.2982 & 6.5271 \\
\hline
\multirow{2}{*}{\textbf{XGboost}}& RMSE & 1.3441 & 1.9853 & 1.9594 \\
                                 & MAPE & 4.7761 & 6.8312 & 6.8653 \\
\hline
\multirow{2}{*}{\textbf{DNN}}& RMSE & 1.4330 & 1.9740 & 3.0581 \\
                                 & MAPE & 5.1147 & 6.7127 & 10.7883 \\
\hline
\multirow{2}{*}{\textbf{Persistent}}& RMSE & 1.1512 & 1.0945 & 1.1537 \\
                                 & MAPE & 4.0757 & 3.7594 & 3.9751 \\
\hline
\end{tabular}
\caption{ EFFECT OF DIFFERENT TIME INTERVALS ON PREDICTION ACCURACY}
\end{table}

\begin{algorithm}[h]
\caption{STFM-V1 Training Procedure}
\label{alg:STFM}
\begin{algorithmic}
\STATE \textbf{Exegesis:} \( K \): set of unique anti-diagonal indices in the matrix; \( n_k \): number of elements per anti-diagonal.
\hrule

\STATE \textbf{while not converged do}
\STATE \quad \textbf{for} \( t = 0, \dots, n_1 \) \textbf{do}

\STATE \quad \quad Sample SST data \( X \in \mathbb{R}^{M \times N}, D \in \mathbb{R}^{L \times M} \)

\STATE \quad \quad \textbf{STEP 1: Obtaining Trend and Seasonal Data}
\STATE \quad \quad \quad \( T = \text{Trend}(X^T), T \in \mathbb{R}^{N \times M} \)
\STATE \quad \quad \quad \( S = \text{Season}(X^T), S \in \mathbb{R}^{N \times M} \)

\STATE \quad \quad \textbf{STEP 2: Prediction of Seasonal and Trend Data }
\STATE \quad \quad \quad \( \text{Linear}(T_{N \times M}) = W_{N \times N} \cdot T_{N \times M} + b = G_{N \times M} \)
\STATE \quad \quad \quad \( \text{Linear}(S_{N \times M}) = W_{N \times N} \cdot S_{N \times M} + b = F_{N \times M} \)
\STATE \quad \quad \quad \( \tilde{X}^T_{N \times M} = G_{N \times M} + F_{N \times M} \)

\STATE \quad \quad \textbf{STEP 3:Mapping $STFM-V1$ of $\tilde{X}$ and delayed attractor }

\STATE \quad \quad \quad \( 
\begin{aligned}
X_{\text{hidden} \times M}^1 = & \text{Dropout}(\text{ReLU}(\text{LayerNorm}(W_{\text{hidden} \times N} \cdot \tilde{X}^T + b))) 
\end{aligned}
\)

\STATE \quad \quad \quad \( \textstyle \fontsize{8}{12}\selectfont \text{Attention\_Output}_{\text{hidden} \times M} = \text{Self-Attention}(X_{\text{hidden} \times M}^1) \)

\STATE \quad \quad \quad \( 
\begin{aligned}
X_{\frac{\text{hidden}}{2} \times M}^2 = & \text{Dropout}(\text{ReLU}(\text{LayerNorm}(W_{\frac{\text{hidden}}{2} \times \text{hidden}} \cdot X^1 + b)))+\text{Attention\_Output}
\end{aligned}
\)

\STATE \quad \quad \quad \( 
\begin{aligned}
X_{\frac{\text{hidden}}{2} \times M}^3 = & \text{Dropout}(\text{ReLU}(\text{LayerNorm}(W_{\frac{\text{hidden}}{2} \times \frac{\text{hidden}}{2}} \cdot X^2 + b))) 
\end{aligned}
\)

\STATE \quad \quad \quad \( 
\begin{aligned}
X_{\frac{\text{hidden}}{2} \times M}^4 = & \text{Dropout}(\text{ReLU}(\text{LayerNorm}(W_{\frac{\text{hidden}}{2} \times \frac{\text{hidden}}{2}} \cdot X^3 + b))) + X^3
\end{aligned}
\)

\STATE \quad \quad \quad \( 
\begin{aligned}
X_{\frac{\text{hidden}}{2} \times M}^5 = & \text{Dropout}(\text{ReLU}(\text{LayerNorm}(W_{\frac{\text{hidden}}{2} \times \frac{\text{hidden}}{2}} \cdot X^4 + b))) 
\end{aligned}
\)

\STATE \quad \quad \quad \( 
\begin{aligned}
X_{\frac{\text{hidden}}{2} \times M}^6 = & \text{Dropout}(\text{ReLU}(\text{LayerNorm}(W_{\frac{\text{hidden}}{2} \times \frac{\text{hidden}}{2}} \cdot X^5 + b)))+ X^5
\end{aligned}
\)

\STATE \quad \quad \quad \( \text{Output}_{L \times M} = (W_{L \times \frac{\text{hidden}}{2}} \cdot X^6 + b) \)

\STATE \quad \quad \textbf{STEP 4: Add Residual \( x_k(t_M) \) and Parameters \( \lambda_{\text{base}} \)}
\STATE \quad \quad \quad \( \text{Re}_{L \times M} = 1_{L \times M} \cdot x_k(t_M) \)
\STATE \quad \quad \quad \( \text{Output'}_{L \times M} = \text{Output}_{L \times M} + \text{Re}_{L \times M} \times \lambda_{\text{base}} \)

\STATE \quad \quad \textbf{STEP 5: Add the Inverse Diagonal Loss Function}

\[
\scriptsize
\text{diagonal\_loss} = \frac{1}{|K|} \sum_{k \in K} \frac{1}{n_k} \sum_{i=1}^{n_k} ( \text{output}_{k,i}-\frac{1}{n_k} \sum_{j=1}^{n_k} \text{output}_{k,j} )^2
\]

\STATE \quad \quad \textbf{STEP 6: Loss Calculation and Optimization}
\STATE \quad \quad \quad \( 
\text{loss}(\Psi(O), D) = \text{MSE}(\text{Output'}_{L \times M}, D) + \lambda_{\text{diag}} \times\text{diagonal\_loss} 
\)

\STATE \quad \quad \quad \( \text{loss.backward();} \)
\STATE \quad \quad \quad \( \text{optimizer.step();} \)

\STATE \quad \textbf{end for}
\STATE \textbf{end while}
\end{algorithmic}
\end{algorithm}

\begin{table}[h]
\centering
\large 
\setlength{\tabcolsep}{6pt}
\renewcommand{\arraystretch}{1.25} 
\begin{tabular}{ccccccc}
\hline
& \textbf{L} & 2 & 8 & 15 \\
\hline
\multirow{2}{*}{\textbf{STFM}} & RMSE & 0.8011 & 0.5579 & 0.7994 \\
                                 & MAPE & 2.3665 & 1.6323 & 2.6460 \\
\hline
\multirow{2}{*}{\textbf{STFM-V1}}   & RMSE & 0.9210 & 0.4531 & 0.6694 \\
                                 & MAPE & 3.2524 & 1.3878 & 2.2319 \\
\hline
\multirow{2}{*}{\textbf{LSTM}}& RMSE & 0.9915 & 1.0438 & 1.8594 \\
                                 & MAPE & 3.3296 & 3.4261 & 6.5268 \\
\hline
\multirow{2}{*}{\textbf{XGboost}}& RMSE & 0.1236 & 1.2728 & 1.9594 \\
                                 & MAPE & 0.3155 & 4.3134 & 6.8653 \\
\hline
\multirow{2}{*}{\textbf{DNN}}& RMSE & 1.3052 & 2.1858 & 3.0581 \\
                                 & MAPE & 4.3710 & 7.5077 & 10.7883 \\
\hline
\multirow{2}{*}{\textbf{Persistent}}& RMSE & 0.4168 & 0.6042 & 1.1537 \\
                                 & MAPE & 1.0555 & 1.8574 & 3.9751 \\
\hline
\end{tabular}
\caption{Effect of different prediction lengths on prediction accuracy}
\end{table}

\subsection*{4.2 Parallel Experiment}

\begin{figure}[htbp]
  \centering
  \includegraphics[width=7in]{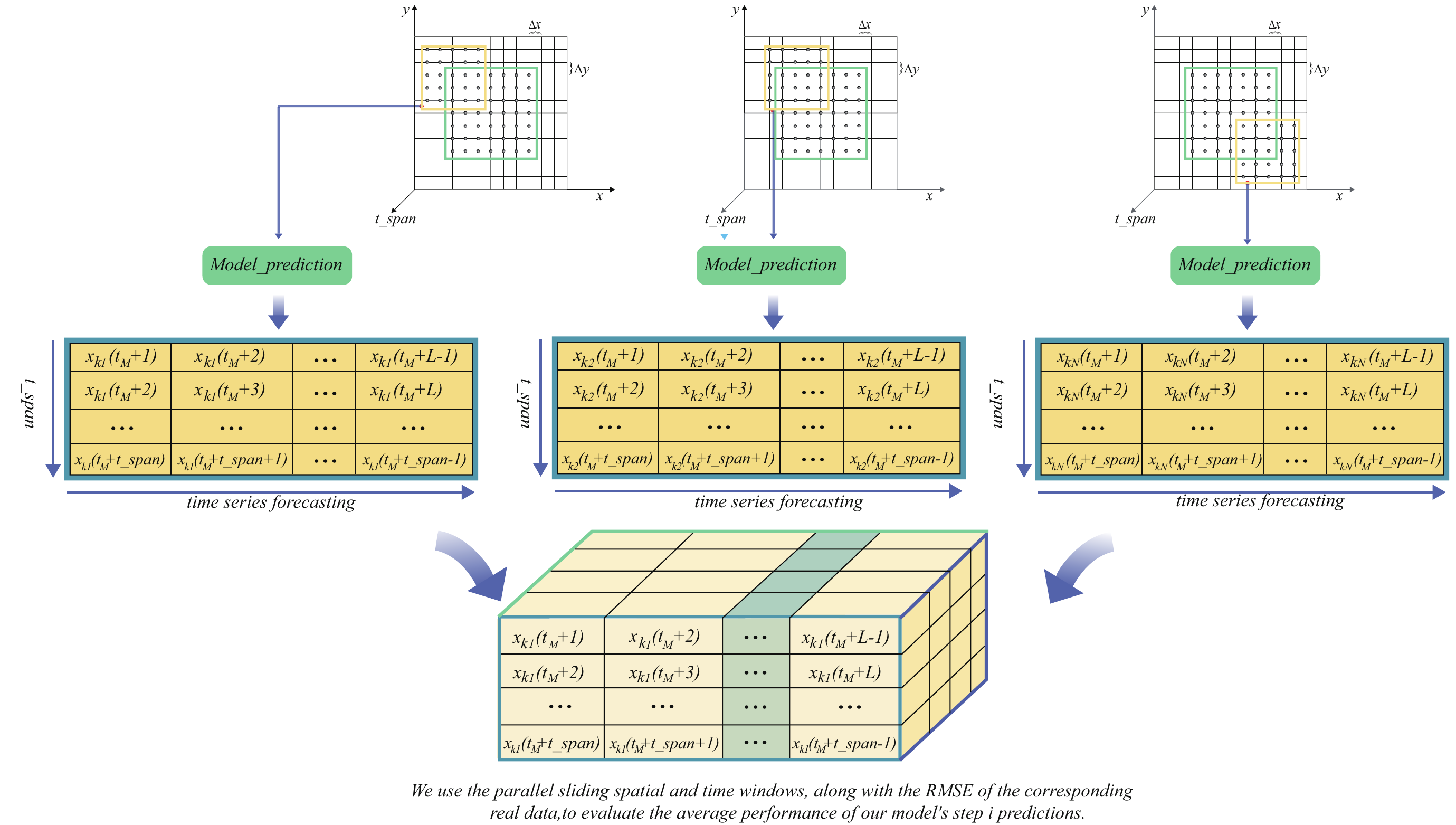} 
  \caption{Model structure for sliding time and space parallelization.} 
  \label{fig:8}  
\end{figure}

From the structure of the STFM-V1 model, the temperature variation at a given point on the sea surface is related to the temperature variations of nearby points, and the prediction process for each point can be considered as an independent process. To test the prediction performance over large ocean areas, we used the sliding sampling method shown in Figure 2 to parallelize predictions spatially. Simultaneously, we performed \( t_{\text{span}} \) time sliding for the same point, and finally, we calculated the RMSE between the predicted values and the corresponding true values at the same prediction step. The overall experimental process is shown in \autoref{fig:8}.

The experiment is based on October 1, 2013, as the reference time, using a fixed time step \( \Delta t = 5 \) to construct prediction sequences, with a prediction period of \( L = 15 \) days, a sliding time span of \( t_{\text{span}} = 10 \) days, and a sampling window length of \( M = 30 \) days. The spatial domain is centered at \( 10.125^\circ \text{N}, 115.125^\circ \text{E} \) with a rectangular primary sampling region of size \( 1^\circ \times 1^\circ \) (see supplementary materials for size parameters). A dynamic correlation analysis window of size \( 2^\circ \times 2^\circ \) is constructed around each grid point in this area, and full-domain prediction is achieved through a sliding mechanism (spatial architecture shown in \autoref{fig:8}).

\begin{figure}[htbp]
  \centering
  \includegraphics[width=4in]{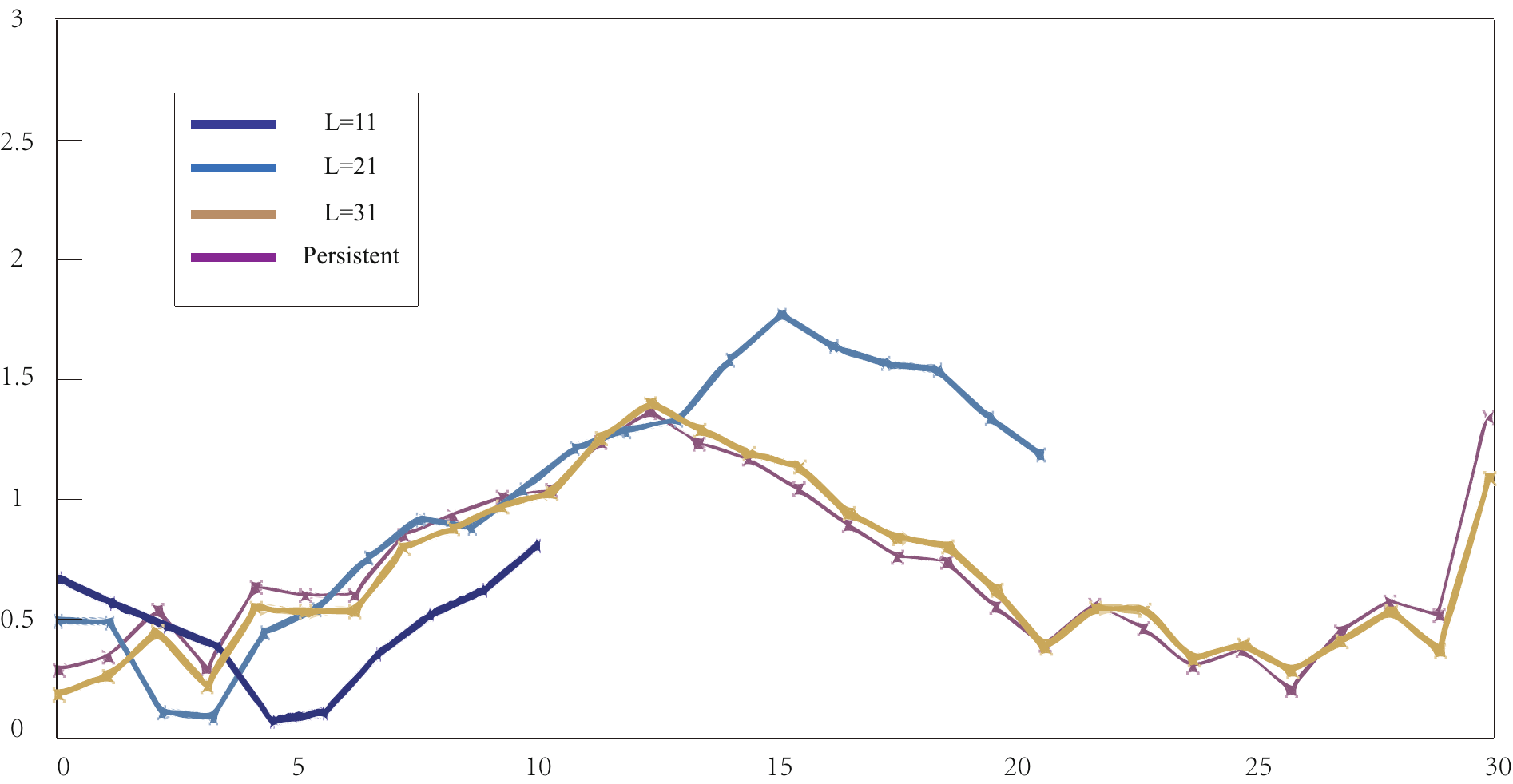} 
  \caption{RMSE of predictions at different \( L \) values.} 
  \label{fig:9}  
\end{figure}

As shown in \autoref{fig:9}, the model performs best in short-term predictions (1-5 steps, 1-25 days), with the RMSE reaching the lowest value. As the prediction period extends to medium-term predictions (6-10 steps, 30-50 days), the system exhibits progressive error accumulation, and the stability decreases. Notably, in the long-term prediction phase (11-15 steps, 55-75 days), the model's performance gradually converges to that of the Persistent Model, ultimately leveling off. This indicates that our model performs reliably in the medium-term (10-step) prediction.

\begin{figure}[htbp]
  \centering
  \includegraphics[width=3.5in]{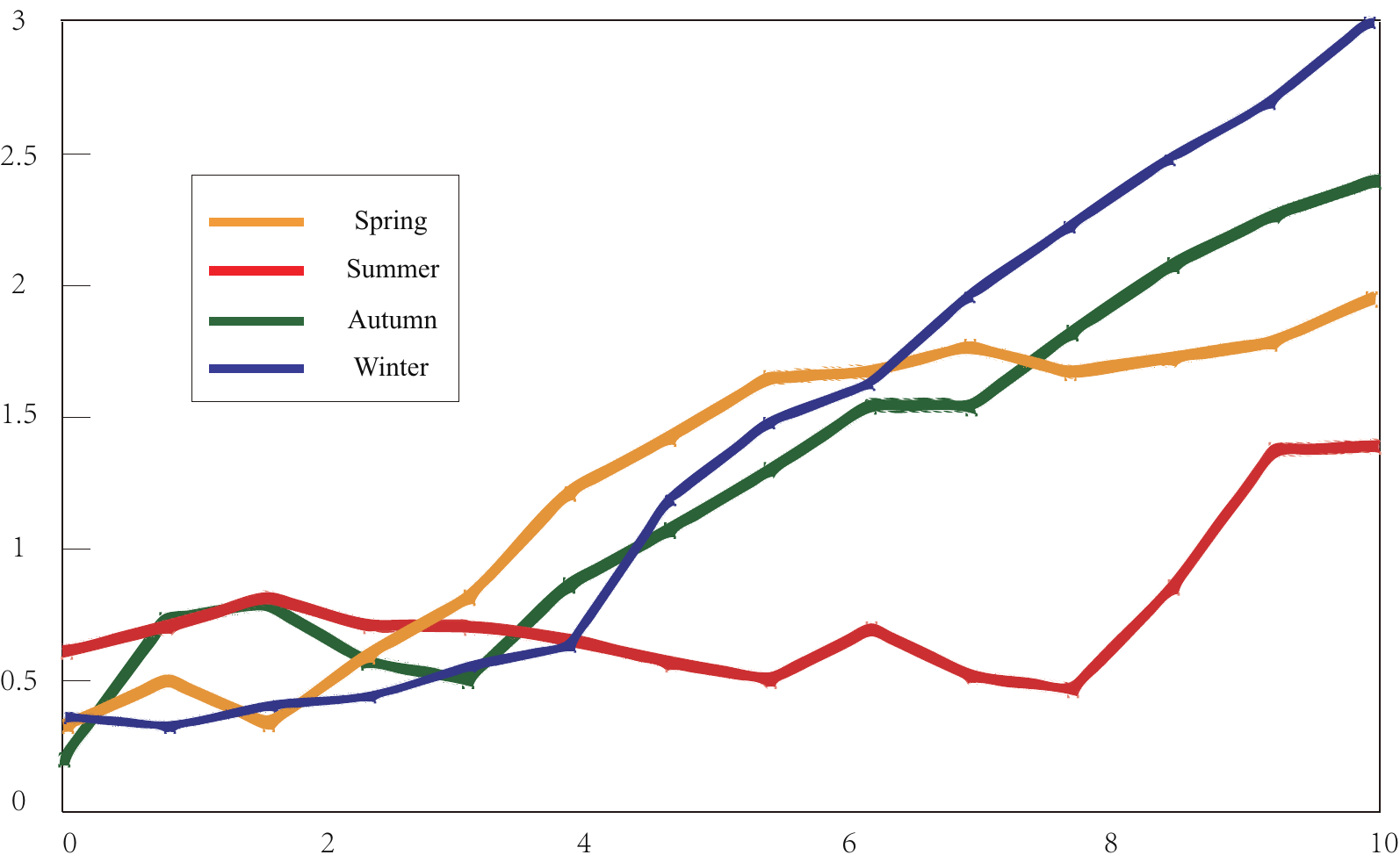} 
  \caption{Prediction RMSE of STFM-V1 in different seasons.}  
  \label{fig:10} 
\end{figure}

\begin{figure}[htbp]
  \centering
  \includegraphics[width=3.5in]{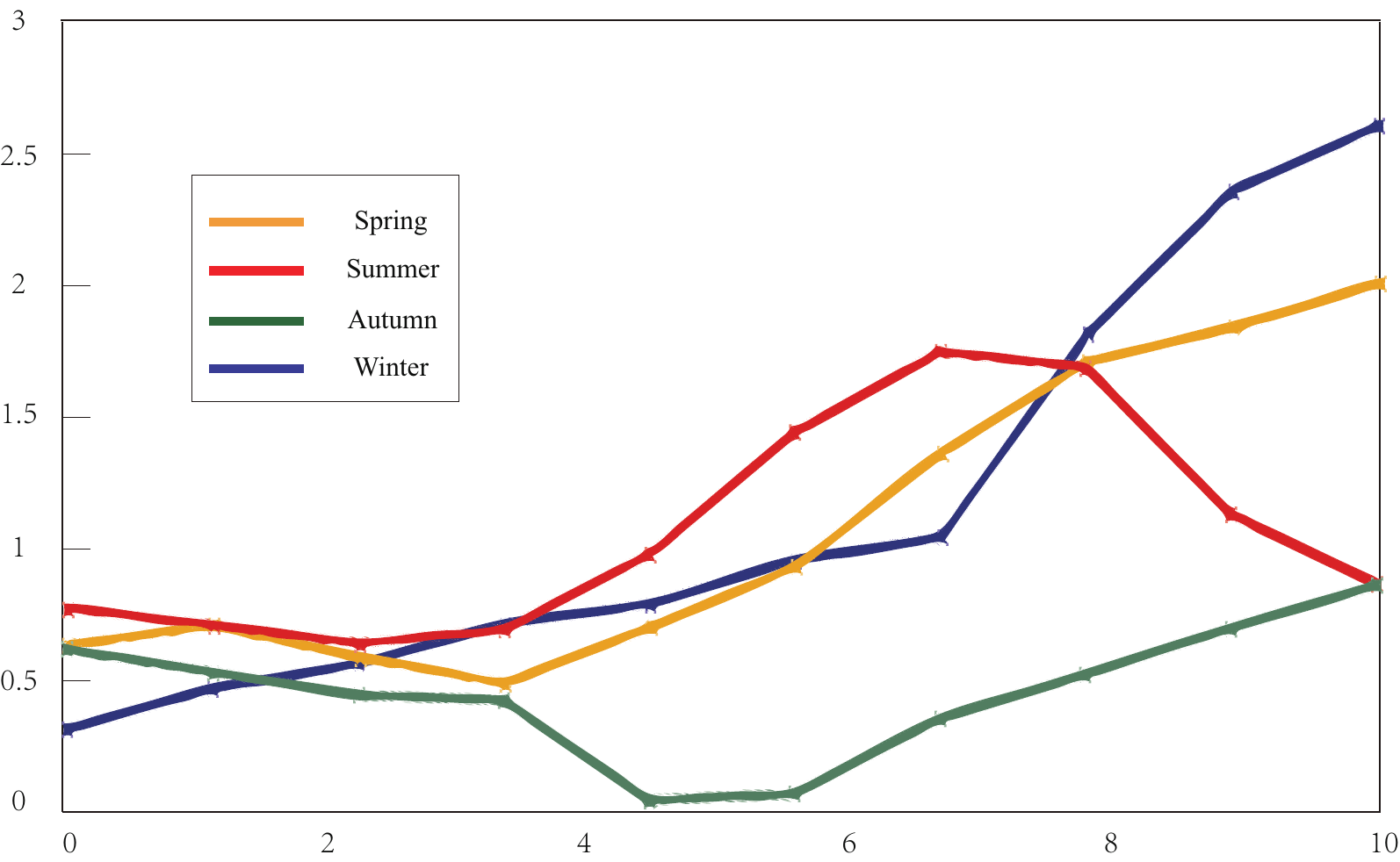}  
  \caption{Prediction RMSE of STFM in different seasons.}  
  \label{fig:11} 
\end{figure}

\begin{figure}[htbp]
  \centering
  \includegraphics[width=3.5in]{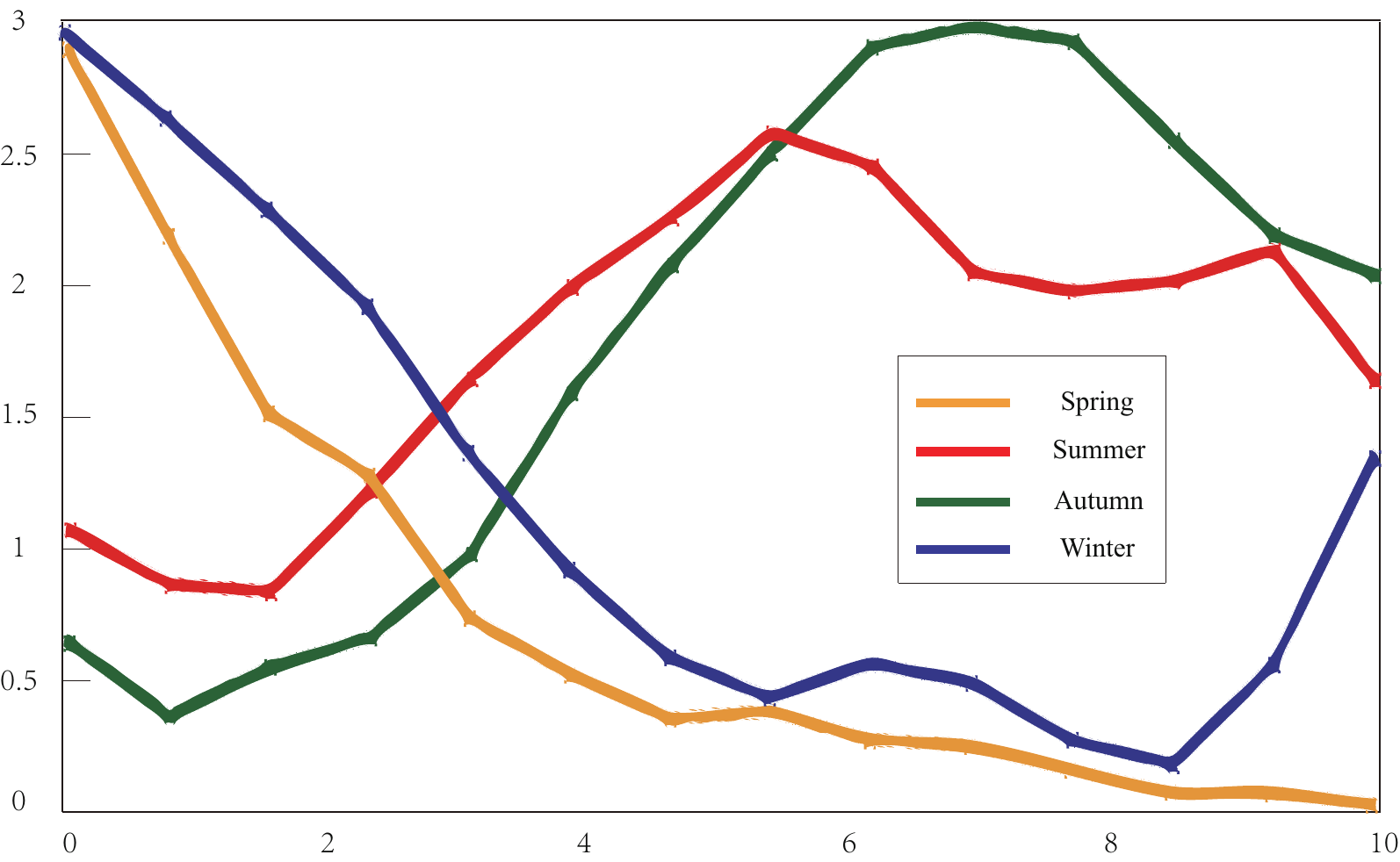} 
  \caption{Prediction RMSE of LSTM/XGBoost in different seasons.} 
  \label{fig:12} 
\end{figure}

\begin{figure}[htbp]
  \centering
  \includegraphics[width=3.5in]{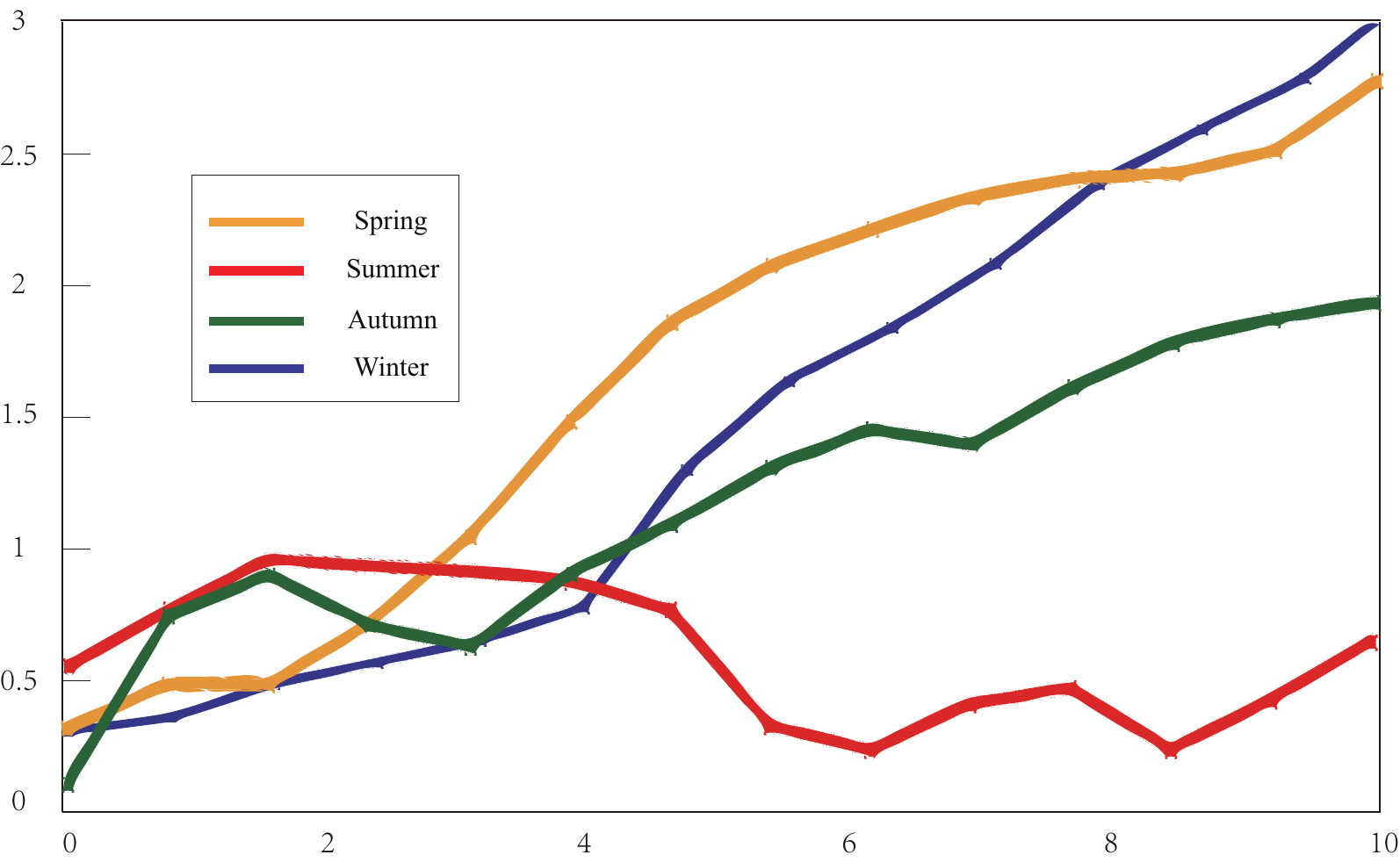} 
  \caption{Prediction RMSE of the Persistent Model in different seasons.} 
  \label{fig:13}  
\end{figure}

Considering the seasonal differences in sea surface temperature (SST) dynamics, we systematically evaluated the prediction performance of five models: STFM-V1 (\autoref{fig:10}), DLNN (\autoref{fig:11}), LSTM (\autoref{fig:12}), XGBoost, and the Persistent Model (\autoref{fig:13}) across the four seasons: spring, summer, autumn, and winter. Notably, LSTM and XGBoost exhibited nearly identical prediction trends, so for simplicity, we only present the results of LSTM.

Comparison analysis reveals a significant pattern of error evolution: the RMSE of STFM-V1, STFM, and the Persistent Model increases as the prediction time extends, while the RMSE of LSTM and XGBoost exhibits a negative correlation with prediction time. This anomalous behavior in purely time-sequential models contradicts the expected positive correlation observed in physically consistent systems, where nodes closer to the observation have higher accuracy. The error reduction in LSTM/XGBoost indicates that these algorithms failed to effectively capture the intrinsic features of the spatiotemporal dynamical system, leading to distorted prediction trends and limited accuracy.

In contrast, the STFM-V1 and STFM models, which integrate spatiotemporal information, show an error growth pattern consistent with the Persistent Model. In terms of prediction accuracy, STFM-V1 significantly outperforms the other models across all seasons. Further analysis reveals that the model performs best in autumn and weakest in winter.

\subsection*{4.3 Ablation Study}

To validate the contributions of different components in our proposed model, we designed an ablation study. By progressively removing or modifying certain components of the model, we evaluated the impact of each part on the final performance, helping to better understand the role of each module.

At the start of the study, we designed a model with the STFM module, trend decomposition, and seasonal decomposition. However, due to the simplicity of the STFM module, the model could not learn deeper features, and there were issues with training instability. As a result, we replaced STFM with STFM-V1, and the comparison showed a significant improvement in model performance.

To explore the impact of trend decomposition and seasonal decomposition on the model, we applied only STFM and STFM-V1 separately and compared their performance. Experimental results indicate that adding trend decomposition and seasonal decomposition slightly improved the model’s performance.

\begin{figure}[htbp]
  \centering
  \includegraphics[width=3.5in]{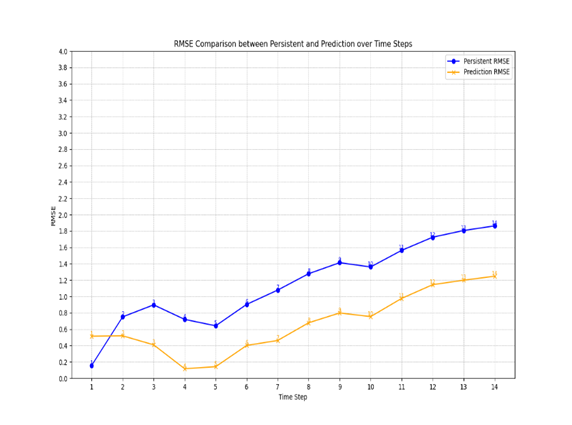} 
  \caption{Comparison of predictions with the Persistent Model.}  
  \label{fig:14}  
\end{figure}

Next, we enhanced the original STFM-V1 model by incorporating diagonal consistency, a self-attention mechanism, and a reference value. Experimental results show that the new model with these three optimizations achieved significant improvements, as shown in \autoref{fig:14}, where its prediction performance outperforms the Persistent Model.

To separately test the contributions of these three optimizations, we removed diagonal consistency, the self-attention mechanism, and the reference value one by one. As shown in Table IV, adding the self-attention mechanism slightly improved the model, while diagonal consistency made a larger contribution to the model’s prediction accuracy. The introduction of the reference value had the greatest contribution to improving the model’s accuracy.

\begin{table}[H]
\centering
\setlength{\tabcolsep}{3.5pt}
\renewcommand{\arraystretch}{1.25}

\begin{tabular}{c|ccccccccc|}
\hline
\multicolumn{1}{c|}{\textbf{Component}} & \multicolumn{8}{c}{\textbf{Choice}} \\ \hline
\textbf{STFM} & \checkmark &  &\checkmark   & &  &  &  &  \\ 
\textbf{STFM-V1} &  & \checkmark &  &\checkmark   &\checkmark   &\checkmark   &\checkmark   & \checkmark \\ 
\textbf{Anti Diagonal Loss} &  &  &  &  &\checkmark  &  & \checkmark &\checkmark  \\ 
\textbf{Self-Attention} &  &  &  &  &\checkmark  & \checkmark &  &\checkmark  \\ 
\textbf{Benchmark Value} &  &  &  &  & \checkmark &\checkmark  &\checkmark  &  \\ 
\textbf{Temporal Decompose} &\checkmark  &\checkmark  &  &  &\checkmark  & \checkmark &\checkmark  & \checkmark \\ 
\textbf{Seasonal Decompose} &\checkmark  &\checkmark  &  &  &\checkmark  &\checkmark  & \checkmark & \checkmark \\ \hline

\textbf{RMSE} & 1.12 & 0.95 & 1.49 & 0.97 & 0.66 & 1.38 & 0.72 & 2.07 \\ 
\textbf{MAPE} & 3.86 & 3.18 & 5.12 & 3.24 & 2.19 & 4.83 & 2.40 & 7.22 \\ \hline
\end{tabular}
\caption{Effect of different components on model prediction accuracy}
\label{tab:prediction_rmse}
\end{table}

\section{Conclusion}

 In this study, we propose a novel model, STFM, based on phase space reconstruction techniques, for precise prediction of Sea Surface Temperature (SST). In the spatiotemporal information fusion module of STFM, by adding different components, we obtained a more stable model — STFM-V1. Through testing its Root Mean Square Error (RMSE) at different variable scales, we validated the high accuracy of the STFM-V1 model in SST prediction. Finally, through ablation experiments, we assessed the contribution of each component to the model's performance.

However, there are still some issues with our model. First, STFM only integrates the temporal and spatial information of the SST data itself, neglecting the influence of external factors such as ocean currents and wind fields on the sea temperature. In future research, we plan to incorporate these factors into the model to improve the accuracy of SST predictions. Second, we observed that STFM's performance varies significantly across different seasons, which may be due to the model's inability to fully learn seasonal changes or the impact of relative positions. To address this, we consider embedding time information as vectors in the time series\cite{church2017word2vec} or embedding relative positions in the time series\cite{lim2021temporal} to further improve the model's seasonal adaptability.

Furthermore, we recognize the potential of integrating advanced self-supervised learning techniques, such as those used in the SLCGC \cite{2025arXiv250203497D} method for hyperspectral image clustering, to improve the model’s ability to extract more robust features from complex spatiotemporal data.

In addition, we plan to apply the model to more fields. Given that phase space reconstruction and STFM-V1 can effectively capture the nonlinear features in existing systems, we hope to extend its application to prediction tasks in fluid dynamic turbulence in the future.

\section*{Acknowledgments}
This was was supported in part by the Shandong Provincial Natural Science Foundation (Grant No. ZR2024QD287) ,Shandong Zhike Intelligence Computing Co., Ltd. and Jinan Fengdi Intelligent Electronics Co., Ltd. .We thank NOAA for providing the 1/4° daily OISST v2.1 dataset used in our method validation.

\bibliographystyle{unsrt}  
\bibliography{STFM}

\end{document}